\documentclass[11pt,a4paper]{article}
\usepackage[hyperref]{emnlp2018}
\usepackage{times}
\usepackage{latexsym}
\usepackage{bm}
\usepackage{amsmath}
\usepackage{graphicx}
\usepackage{amsfonts}
\usepackage{color}
\usepackage{enumitem}

\usepackage{url}

\DeclareMathOperator*{\argmax}{arg\,max}
\aclfinalcopy 

\title{Neural Cross-Lingual Named Entity Recognition with Minimal Resources}

\author{
  Jiateng Xie,$^1$ Zhilin Yang,$^1$Graham Neubig,$^1$ \\\textbf{Noah A. Smith,$^{2,3}$ and Jaime Carbonell$^1$}
 \\
$^1$Language Technologies Institute, Carnegie Mellon University\\
$^2$Paul G. Allen School of Computer Science \& Engineering, University of Washington\\
$^3$Allen Institute for Artificial Intelligence\\
  {\tt \{jiatengx,zhiliny,gneubig,jgc\}@cs.cmu.edu},\\{\tt nasmith@cs.washington.edu} \\ }
\date{}

\begin{document}
\maketitle
\begin{abstract}
For languages with no annotated resources, unsupervised transfer of natural language processing models such as named-entity recognition (NER) from resource-rich languages would be an appealing capability.
However, differences in  words and word order across languages make it a challenging problem.
To improve mapping of lexical items across languages, we propose a method that finds translations based on bilingual word embeddings.
To improve robustness to word order differences, we propose to use self-attention, which allows for a degree of flexibility with respect to word order.
We demonstrate that these methods achieve state-of-the-art or competitive NER performance on commonly tested languages under a cross-lingual setting, with much lower resource requirements than past approaches.
We also evaluate the challenges of applying these methods to Uyghur, a low-resource language.\footnote{The source code is available at \url{https://github.com/thespectrewithin/cross-lingual_NER}}
\end{abstract}

\section{Introduction}
\label{intro}

Named entity recognition (NER), the task of detecting and classifying named entities from text into a few predefined categories such as people, locations or organizations, has seen the state-of-the-art greatly advanced by the introduction of neural architectures ~\citep{collobert2011natural,DBLP:journals/corr/HuangXY15,DBLP:journals/corr/ChiuN15,DBLP:conf/naacl/LampleBSKD16,DBLP:journals/corr/YangSC16,ma-hovy:2016:P16-1,P17-1161,DBLP:journals/corr/abs-1709-04109,Peters:2018}.
However, the success of these methods is highly dependent on a reasonably large amount of annotated training data, and thus it remains a challenge to apply these models to languages with limited amounts of labeled data. Cross-lingual NER attempts to address this challenge by transferring knowledge from a high-resource source language with abundant entity labels to a low-resource target language with few or no labels.
Specifically, in this paper we attempt to tackle the extreme scenario of \emph{unsupervised transfer}, where no labeled data is available in the target language.
Within this paradigm, there are two major challenges to tackle: how to effectively perform lexical mapping between the languages, and how to address word order differences.


To cope with the first challenge of lexical mapping, a number of methods use parallel corpora to project annotations between languages through word alignment~\citep{ehrmann2011building,kim2012multilingual,Wang-Manning:tacl14:cross,DBLP:journals/corr/NiDF17}.
Since parallel corpora may not be always available, ~\citet{D17-1268} proposed a ``cheap translation'' approach that uses a bilingual dictionary to perform word-level translation.
The above approaches provide a reasonable proxy for the actual labeled training data, largely because the words that participate in entities can be translated relatively reliably given extensive parallel dictionaries or corpora (e.g., with 1 million word pairs or sentences).
Additionally, as a side benefit of having explicitly translated words, models can directly exploit features extracted from the surface forms (e.g. through character-level neural feature extractors), which has proven essential for high accuracy in the monolingual scenario~\cite{ma-hovy:2016:P16-1}.
However, these methods are largely predicated on the availability of large-scale parallel resources, and thus, their applicability to low-resource languages is limited.



\begin{figure*}[th]
\centering
\includegraphics[width=\textwidth]{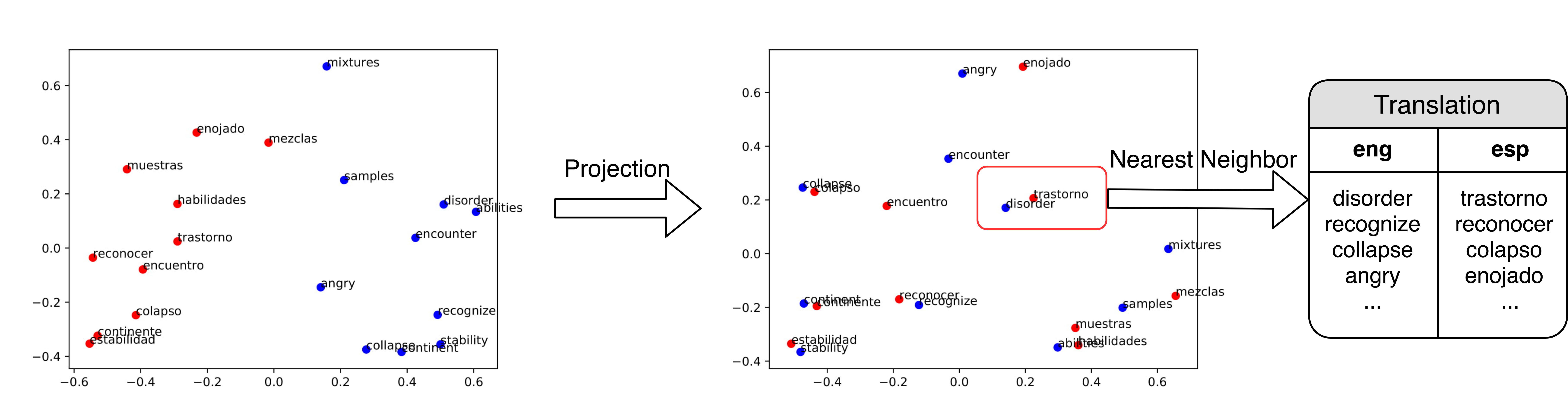}
\caption{Example of the result of our approach on Spanish-English words not included in the dictionary (embeddings are reduced to 2 dimensions for visual clarity). We first project word embeddings into a shared space, and then use the nearest neighbors for word translation. Notice that the word pairs are not perfectly aligned in the shared embedding space, but after word translation we obtain correct alignments.}
\label{bwet}
\end{figure*}

In contrast, it is also possible to learn lexical mappings through bilingual word embeddings (BWE).
These bilingual embeddings can be obtained by using a small dictionary to project two sets of embeddings into a consistent space~\citep{DBLP:journals/corr/MikolovLS13,faruqui2014improving,artetxe2016learning,DBLP:journals/corr/SmithTHH17},
or even in an entirely unsupervised manner using adversarial training or identical character strings ~\citep{zhang2017adversarial,artetxe2017learning,lample2018word}. Many approaches in the past have leveraged the shared embedding space for cross-lingual applications~\citep{guo2015cross,ammar:2016:massively,DBLP:conf/naacl/ZhangGBJ16,fang-cohn:2017:Short}, including NER~\citep{D16-1153,DBLP:journals/corr/NiDF17}.
The minimal dependency on parallel resources makes the embedding-based method much more suitable for low-resource languages.
However, since different languages have different linguistic properties, it is hard, if not impossible, to align the two embedding spaces perfectly (see Figure \ref{bwet}). 
Meanwhile, because surface forms are not available, character-level features cannot be used, resulting in reduced tagging accuracy (as demonstrated in our experiments).


To address the above issues, we propose a new lexical mapping approach that combines the advantages of both discrete dictionary-based methods and continuous embedding-based methods.
Specifically, we first project embeddings of different languages into the shared BWE space, then learn discrete word translations by looking for nearest neighbors in this projected space, and finally train a model on the translated data. This allows our method to inherit the benefits of both embedding-based and dictionary-based methods: its resource requirements are low as in the former, but it suffers less from misalignment of the em- bedding spaces and has access to character-level information like the latter. 


Turning to differences in word ordering, to our knowledge there are no methods that explicitly deal with this problem in unsupervised cross-lingual transfer for NER.
Our second contribution is a method to alleviate this issue by incorporating an order-invariant self-attention mechanism~\citep{NIPS2017_7181,lin+al-2017-embed-iclr} into our neural architecture.
Self-attention allows re-ordering of information within a particular encoded sequence, which makes it possible to account for word order differences between the source and the target languages.


In our experiments, we start with models trained in English as the source language on the CoNLL 2002 and 2003 datasets and transfer them into Spanish, Dutch, and German as the target languages.
Our approach obtains new state-of-the-art cross-lingual results in Spanish and Dutch, and competitive results in German, even without a dictionary, completely removing the need for resources such as Wikipedia and parallel corpora. 
Next, we transfer English using the same approach into Uyghur, a truly low-resource language. 
With significantly fewer cross-lingual resources, our approach can still perform competitively with previous best results.

\section{Approach}
\label{bwe}

We establish our problem setting (\S\ref{sec:problem}), then present our methods in detail (\S\ref{sec:method}), and provide some additional motivation (\S\ref{sec:disc}).

\subsection{Problem Setting} \label{sec:problem}

NER takes a sentence as the input and outputs a sequence of labels corresponding to the named entity categories of the words in the sentence, such as location, organization, person, or none. 
In standard supervised NER, we are provided with a labeled corpus of sentences in the target language along with tags indicating which spans correspond to entities of each type.

As noted in the introduction, we study the problem of unsupervised cross-lingual NER:  given labeled training data only in a separate source language, we aim to learn a model that is able to perform NER in the target language.
This transfer can be performed using a variety of resources, including parallel corpora \cite{Tackstrom:2012,DBLP:journals/corr/NiDF17}, Wikipedia \cite{nothman2013learning}, and large dictionaries \cite{DBLP:journals/corr/NiDF17,D17-1268}.
In this work, we limit ourselves to a setting where we have the following resources, making us comparable to other methods such as \citet{D17-1268} and \citet{DBLP:journals/corr/NiDF17}:
\begin{itemize}[leftmargin=1.0em,topsep=0.5em,itemsep=0em]
\item Labeled training data in the source language.
\item Monolingual corpora in both source and target languages.
\item A dictionary, either a small pre-existing one, or one induced by unsupervised methods.
\end{itemize}


\subsection{Method} \label{sec:method}

Our method follows the process below:
\begin{enumerate}[leftmargin=1.0em,topsep=0.5em,itemsep=0em]
\item Train separate word embeddings using monolingual corpora using standard embedding training methods (\S\ref{sec:mwe}).
\item Project word embeddings in the two languages into a shared embedding space by optimizing the word embedding alignment using the given dictionary (\S\ref{sec:bwe}).
\item For each word in the source language training data, translate it by finding its nearest neighbor in the shared embedding space (\S\ref{sec:trans}). 
\item Train an NER model using the translated words along with the named entity tags from the English corpus (\S\ref{sec:sup}).
\end{enumerate}
We consider each in detail.

\subsubsection{Learning Monolingual Embeddings} \label{sec:mwe}

Given text in the source and target language, we first independently learn word embedding matrices $X$ and $Y$ in the source and target languages respectively.
These embeddings can be learned on monolingual text in both languages with any of the myriad of word embedding methods \cite{NIPS2013_5021,pennington2014glove,bojanowski2017enriching}.

\subsubsection{Learning Bilingual Embeddings} \label{sec:bwe}

Next, we learn a cross-lingual projection of $X$ and $Y$ into a shared space.
Assume we are given a dictionary $\{x_i, y_i\}_{i=1}^D$, where $x_i$ and $y_i$ denote the embeddings of a word pair. Let $X_D = [x_1, x_2, \cdots, x_D]^\top$ and $Y_D = [y_1, y_2, \cdots, y_D]^\top$ denote two embedding matrices consisting of word pairs from the dictionary.




Following previous work~\citep{DBLP:conf/naacl/ZhangGBJ16,artetxe2016learning,DBLP:journals/corr/SmithTHH17}, we optimize the following objective:
\[
\min_W \sum_{i=1}^d \|Wx_i - y_i\|^2 \text{~~s.t.~~} W W^\top = I,
\]
where $W$ is a square parameter matrix. This objective can be further simplified as
\[
\max_W \text{Tr}(X_D W Y_D^\top) \text{~~s.t.~~} W W^\top = I.
\]
Here, the transformation matrix $W$ is constrained to be orthogonal so that the dot product similarity of words is invariant with respect to the transformation both within and across languages.

To optimize the above objective (the Procrustes problem), we decompose the matrix $Y_D^\top X_D$ using singular value decomposition. Let the results be $Y_D^\top X_D = U\sum V^\top$, then $W = U V^\top$ gives the exact solution. We define the similarity matrix between $X$ and $Y$ to be $S = YWX^\top = YU(XV)^\top$, where each column contains the cosine similarity between source word $x_i$ and all target words $y_i$.
We can then define $X'=XV$ and $Y'=YU$, which are $X$ and $Y$ transformed into a shared embedding space.

To refine the alignment in this shared space further, we iteratively perform a self-learning refinement step $k$~\footnote{We use $k=3$ in this paper.} times by:
\begin{enumerate}[leftmargin=1.0em,topsep=0.5em,itemsep=0em]
\item Using the aligned embeddings  to generate a new dictionary that consists of mutual nearest neighbors obtained using the same metric as introduced below.
\item Solving the Procrustes problem based on the newly generated dictionary to get a new set of bilingual embeddings.
\end{enumerate}
The bilingual embeddings at the end of the $k$th step, $X'_{k}$ and $Y'_{k}$, will be used to perform translation.

\subsubsection{Learning Word Translations} \label{sec:trans}

To learn actual word translations, we next proceed to perform nearest-neighbor search in the common space.
Instead of using a common distance metric such as cosine similarity, we adopt the cross-domain similarity local scaling (CSLS) metric ~\citep{lample2018word}, which is designed to address the hubness problem common to the shared embedding space ~\citep{DBLP:journals/corr/DinuB14}. Specifically, 
\[\text{CSLS}(x_i, y_j) = 2\cos(x_i, y_j) - r_T(x_i) - r_S(y_j)
\]
where $r_T(x_i) = \frac{1}{K} \sum_{y_t \in N_T(x_i)} \cos(x_i, y_t)$ denotes the mean cosine similarity between $x_i$ and its $K$ neighbors $y_t$.
Using this metric, we find translations for each source word $s$ by selecting target word $\hat{t_s}$ where $\displaystyle \hat{t_s} = \argmax_t{\text{CSLS}(x_s, y_t)}$.

\subsubsection{Training the NER Model} \label{sec:sup}

Finally, we translate the entire English NER training data into the target language by taking English sentences $S=s_1, s_2, ...,s_n$ and translating them into target sentences $\hat{T}=\hat{t_1}, \hat{t_2}, ..., \hat{t_n}$. The label of each English word is copied to be the label of the target word. We can then train an NER model directly using the translated data. Notably, because the model has access to the surface forms of the target sentences, it can use the character sequences of the target language as part of its input.

During learning, all word embeddings are normalized to lie on the unit ball, allowing every training pair an equal contribution to the objective and improving word translation accuracy \citep{artetxe2016learning}.  When training the NER model, however, we do not normalize the word embeddings, because preliminary experiments showed the original unnormalized embeddings gave superior results.  We suspect this is due to frequency information conveyed by vector length, an important signal for NER.  (Named entities appear less frequently in the monolingual corpus.)


\subsection{Discussion} \label{sec:disc}

Figure \ref{bwet} shows an example of the embeddings and translations learned with our approach trained on Spanish and English data from the experiments (see  \S\ref{exp} for more details).
As shown in the figure, there is usually a noticeable difference between the word embeddings of a word pair in different languages, which is inevitable because different languages have distinct traits and different monolingual data, and as a result it is intrinsically hard to learn a perfect alignment.
This indicates that models trained directly on data using the source embeddings may not generalize well to the slightly different embeddings of the target language.

Instead of directly modeling the shared embedding space~\citep{guo2015cross,DBLP:conf/naacl/ZhangGBJ16,fang-cohn:2017:Short,DBLP:journals/corr/NiDF17}, we leverage the shared embedding space for word translation.
As shown in Figure \ref{bwet}, unaligned word pairs can still be translated correctly with our method, as the embeddings are still closer to the correct translations than the closest incorrect one.

\section{NER Model Architecture}
\label{model}

\begin{figure}[tb]
\centering
\includegraphics[width=0.9\columnwidth,height=10cm]{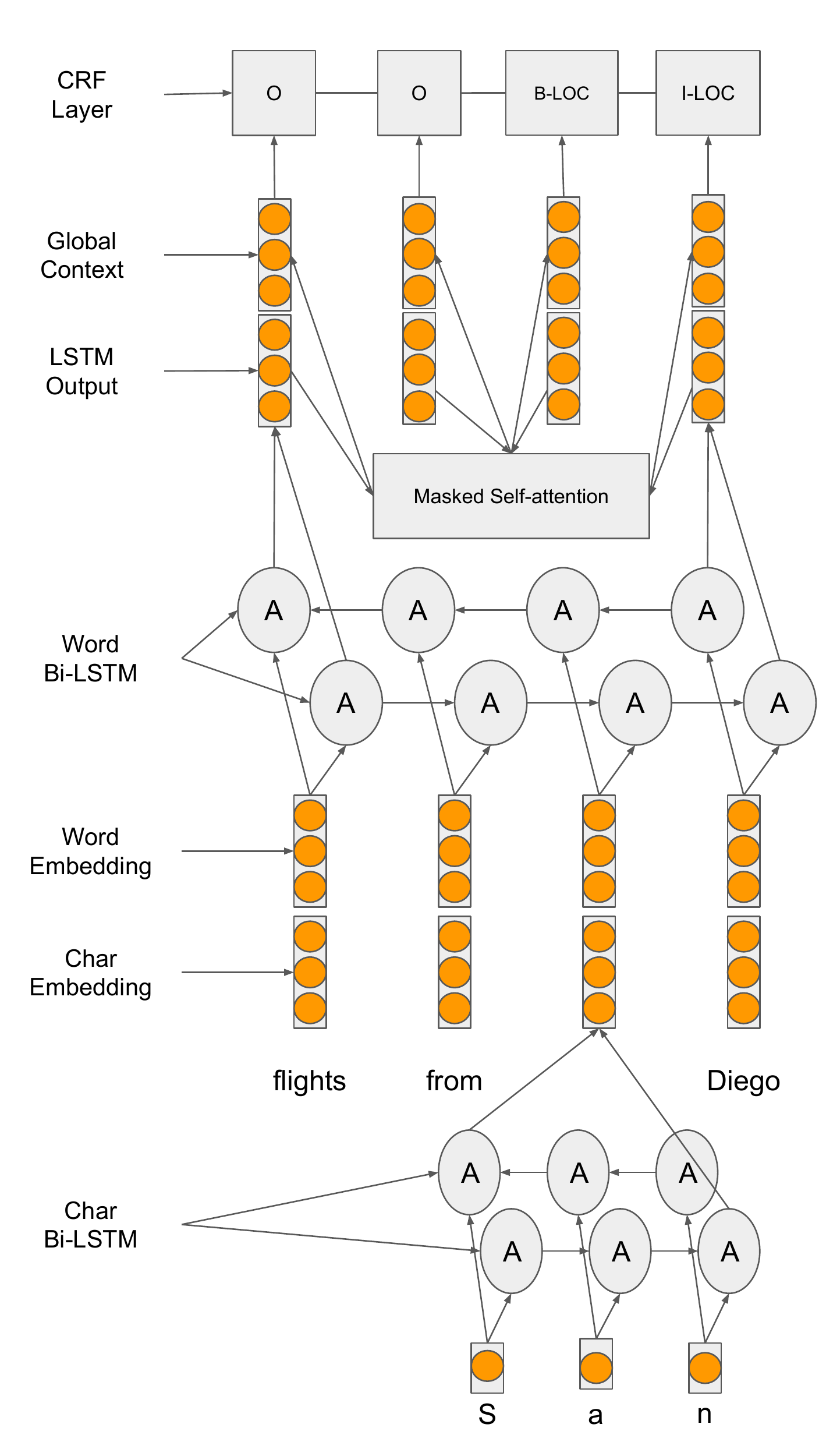}
\caption{Self-attentive Bi-LSTM-CRF Model}
\label{fig:b}
\end{figure}

We describe the model we use to perform NER. We will first describe the basic hierarchical neural CRF tagging model~\citep{DBLP:conf/naacl/LampleBSKD16,ma-hovy:2016:P16-1,DBLP:journals/corr/YangSC16}, and introduce the self-attention mechanism that we propose to deal with divergence of word order.

\subsection{Hierarchical Neural CRF}

The hierarchical CRF model consists of three components: a character-level neural network, either an RNN or a CNN, that allows the model to capture subword information, such as morphological variations and capitalization patterns; a word-level neural network, usually an RNN, that consumes word representations and produces context sensitive hidden representations for each word; and a linear-chain CRF layer that models the dependency between labels and performs inference.



In this paper, we closely follow the architecture proposed by ~\citet{DBLP:conf/naacl/LampleBSKD16}, and use bi-directional LSTMs for both the character level and word level neural networks.
Specifically, given an input sequence of words $(w_1, w_2, ... , w_n)$, and each word's corresponding character sequence, the model first produces a representation for each word, $x_i$, by concatenating its character representation with its word embedding.
Subsequently, the word representations of the input sequence $(x_1, x_2, \cdots, x_n)$ are fed into a word level Bi-LSTM, which models the contextual dependency within each sentence and outputs a sequence of context sensitive hidden representations $(h_1, h_2, \cdots, h_n)$. A CRF layer is then applied on top of the word level LSTM and takes in as its input the sequence of hidden representations $(h_1, h_2, \cdots, h_n)$, and defines the joint distribution of all possible output label sequences. The Viterbi algorithm is used during decoding.

\subsection{Self-Attention}

The training-time inputs to our model are in essence corrupted sentences from the target language (e.g., Spanish), which have a different order from natural target sentences. We propose to alleviate this problem by adding a self-attention layer~\citep{NIPS2017_7181} on top of the word-level Bi-LSTM.
Self-attention provides each word with a context feature vector based on \emph{all} the words of a sentence. As the context vectors are obtained irrespective of the words' positions in a sentence, at test time, the model is more likely to see vectors similar to those seen at training time, which we posit introduces a level of flexibility with respect to the word order, and thus may allow for better generalization. 

Let $H = [h_1, h_2, \cdots, h_n]^\top$ be a sequence of word-level hidden representations. We apply a single layer MLP on $H$ to obtain the queries $Q$ and keys $K = \text{tanh}(H W + b)$, where $W \in \mathbb{R}^{d \times d}$ is a parameter matrix and $b \in \mathbb{R}^d$ is a bias term, with $d$ being the hidden state size. The output of attention layer is defined as:
\begin{align*}
H^a &= \text{softmax}(QK^\top) \odot (E-I) H\\
    &= [h^a_1, h^a_2, ..., h^a_3]^\top
\end{align*}
where $I$ is an identity matrix and $E$ is an all-one matrix. The term $(E-I)$ serves as an attention mask that prevents the weights from centering on the word itself, as we would like to provide each word with sentence level context. The outputs from the self-attention layer are then concatenated with the original hidden representations to form the final inputs to the CRF layer, which are $([h_1, h^a_1], [h_2, h^a_2], ... , [h_3, h^a_3])$.

\section{Experiments}
\label{exp}




To examine the effectiveness of both of our proposed methods, we conduct four sets of experiments. First, we evaluate our model both with and without provided dictionaries on a benchmark NER dataset and compare with previous state-of-the-art results. Second, we compare our methods against a recently proposed dictionary-based translation baseline~\citep{D17-1268} by directly applying our model on their translated data.\footnote{We thank the authors of ~\citet{D17-1268} for sharing their data.} Subsequently, we conduct an ablation study to further understand our proposed methods. Lastly, we apply our methods to a truly low-resource language, Uyghur.
\subsection{Experimental Settings}

We evaluate our proposed methods on the benchmark CoNLL 2002 and 2003 NER datasets ~\citep{TjongKimSang:2002:ICS:1118853.1118877,tjong2003introduction}, which contain 4 European languages, English, German, Dutch and Spanish. 
For all experiments, we use English as the source language and translate its training data into the target language. We train a model on the translated data, and test it on the target language. 
For each experiment, we run our models 5 times using different seeds and report the mean and standard deviation, as suggested by~\citet{Reimers2017ReportingSD}.


\textbf{Word Embeddings}
For all languages, we use two different embedding methods, fastText ~\citep{bojanowski2017enriching} and GloVe ~\citep{pennington2014glove}, to perform word-embedding based translations and train the NER model, respectively. For fastText, we use the publicly available embeddings trained on Wikipedia for all languages. For GloVe, we use the publicly available embeddings pre-trained on Gigaword and Wikipedia for English. For Spanish, German and Dutch, we use Spanish Gigaword and Wikipedia, German WMT News Crawl data and Wikipedia, and Dutch Wikipedia, respectively, to train the GloVe word embeddings.
We use a vocabulary size of 100,000 for both embedding methods.



\textbf{Dictionary}
We consider three different settings to obtain the seed dictionary, including two methods that do not use parallel resources:
\begin{enumerate}[leftmargin=1.0em,topsep=0.5em,itemsep=0em]
\item Use identical character strings shared between the two vocabularies as the seed dictionary. 
\item \citet{lample2018word}'s method of using adversarial learning to induce a mapping that aligns the two embedding spaces, and the mutual nearest neighbors in the shared space will be used as a dictionary. The learning procedure is formulated as a two player game, where a discriminator is trained to distinguish words from the two embedding spaces, and a linear mapping is trained to align the two embedding spaces and thus fool the discriminator.
\item Use a provided dictionary. In our experiments, we use the ones provided by \citet{lample2018word},\footnote{\url{https://github.com/facebookresearch/MUSE}} each of which contain 5,000 source words and about 10,000 entries.
\end{enumerate}


\textbf{Translation}
We follow the general procedure described in Section~\ref{bwe}, and replace each word from the English training data with its corresponding word in the target language. For out-of-vocabulary (OOV) words, we simply keep them as-is. We capitalize the resulting sentences following the pattern of the original English words. Note that for German, simply following the English capitalization pattern does not work, because all nouns in German are capitalized. To handle this problem, we count the number of times each word is capitalized in Wikipedia, and capitalize the word if the probability is greater than $0.6$. 

\textbf{Network Parameters}
For our experiments, we set the character embedding size to be 25, character level LSTM hidden size to be 50, and word level LSTM hidden size to be 200. For OOV words, we initialize an unknown embedding by uniformly sampling from range $[-\sqrt[]{\frac{3}{\text{emb}}}, +\sqrt[]{\frac{3}{\text{emb}}}]$, where emb is the size of embedding, 100 in our case. We replace each number with 0 when used as input to the character level Bi-LSTM.

\textbf{Network Training}
We use SGD with momentum to train the NER model for 30 epochs, and select the best model on the target language development set. We choose the initial learning rate to be $\eta_0 = 0.015$, and update it using a learning decay mechanism after each epoch, $\eta_t = \frac{\eta_0}{1 + \rho t}$, where $t$ is the number of completed epoch and $\rho = 0.05$ is the decay rate. We use a batch size of 10 and evaluate the model per 150 batches within each epoch. We apply dropout on the inputs to the word-level Bi-LSTM, the outputs of the word-level Bi-LSTM, and the outputs of the self-attention layer to prevent overfitting. The self-attention dropout rate is set to 0.5 when using our translated data, and 0.2 when using cheap-translation data. We use 0.5 for all other dropouts. The word embeddings are not fine-tuned during training.

\subsection{Results}

\begin{table*}[!h]
  \begin{center}
  \resizebox{\textwidth}{!}{%
    \begin{tabular}{rl|l|l|l|l}
      \multicolumn{2}{l}{{Model}} & Spanish & Dutch & German & Extra Resources\\
      \hline
      $^*$ & \citet{Tackstrom:2012} & $59.30$ & $58.40$ & $40.40$&parallel corpus\\
      $^*$ & \citet{nothman2013learning} & $61.0$ & $64.00$ & $55.80$&Wikipedia\\
      $^*$ & \citet{TsaiMaRo16} & $60.55$ & $61.60$& $48.10$& Wikipedia\\
      $^*$ & \citet{DBLP:journals/corr/NiDF17} & $65.10$ & $65.40$ & $58.50$& Wikipedia, parallel corpus, 5K dict.\\
      \hline
      $^{*\dagger}$ & \citet{D17-1268} & $65.95$ & $66.50$ & $\mathbf{59.11}$ & Wikipedia, 1M dict.\\
      $^*$ & \citet{D17-1268} (only Eng.~data)& $51.82$ & 53.94 & 50.96 & 1M dict.\\
           \hline
      \multicolumn{2}{l|}{\emph{Our methods:}} &&&&\\
      & BWET (id.c.) & $71.14 \pm 0.60$ & $70.24 \pm 1.18$ & $57.03 \pm 0.25$ & --\\
      & BWET (id.c.) + self-att. & $\mathbf{72.37} \pm 0.65$& $70.40 \pm 1.16$&$\mathbf{57.76} \pm 0.12$ & --\\
      &  BWET (adv.) & $70.54 \pm 0.85$ & $70.13 \pm 1.04$ & $55.71 \pm 0.47$ & --\\
      & BWET (adv.) + self-att. &$71.03 \pm 0.44$ & $\mathbf{71.25} \pm 0.79$& $56.90 \pm 0.76$& --\\
      & BWET & $71.33 \pm 1.26$ & $69.39 \pm 0.53$ & $56.95 \pm 1.20$ & 10K dict.\\
      & BWET + self-att. & $71.67 \pm 0.86$ & $70.90 \pm 1.09$ & $57.43 \pm 0.95$ & 10K dict.\\
            $^*$ & BWET on data from \citet{D17-1268} & $66.53 \pm 1.12$ & $69.24 \pm 0.66$ & $55.39 \pm 0.98$ & 1M dict.\\
      $^*$ & BWET + self-att. on data from \citet{D17-1268} & $66.90 \pm 0.65$ &$69.31 \pm 0.49$ & $55.98 \pm 0.65$ & 1M dict.\\
      \hline
	  $^*$ & Our supervised results & $86.26 \pm 0.40$& $86.40 \pm 0.17$& $78.16 \pm 0.45$& annotated corpus\\
    \end{tabular}
    }
    \caption{NER $F_1$ scores. $^*$Approaches that use more resources than ours (``Wikipedia'' means Wikipedia is used not as a monolingual corpus, but to provide external knowledge). $^\dagger$Approaches that use multiple languages for transfer. ``Only Eng.~data'' is the model used in~\citet{D17-1268} trained on their data translated from English without using Wikipedia and other languages.  The ``data from~\citet{D17-1268}'' is the same data translated from only English they used. ``Id.c.'' indicates using identical character strings between the two languages as the seed dictionary. ``Adv.'' indicates using adversarial training and mutual nearest neighbors to induce a seed dictionary.  Our supervised results are obtained using models trained on annotated corpus from CoNLL.}
    \label{tab:table2}
  \end{center}
  \vspace{-0.5em}
\end{table*}


Table~\ref{tab:table2} presents our results on transferring from English to three other  languages, alongside results from previous studies. Here ``BWET'' (bilingual word embedding translation) denotes using the  hierarchical neural CRF model trained on data translated from English. As can be seen from the table, our  methods outperform previous state-of-the-art results on Spanish and Dutch by a large margin and perform competitively on German even without using any parallel resources. We achieve similar results using different seed dictionaries, and produce the best results when adding the self-attention mechanism to our model.



Despite the good performance on Spanish and Dutch, our model does not outperform the previous best result on German, and we speculate that there are a few reasons. First, German has rich morphology and contains many compound words, making the word embeddings less reliable. 
Our supervised result on German indicates the same problem, as it is about 8 $F_1$ points worse than Spanish and Dutch.
Second, these difficulties become more pronounced in the cross-lingual setting, leading to a noisier embedding space alignment, which lowers the quality of BWE-based translation. We believe that this is a problem with all methods using word embeddings. In such cases, more resource-intensive methods may be necessary.

\subsubsection{Comparison with Dictionary-Based Translation}

Table~\ref{tab:table2} also presents results of a comparison between our proposed BWE translation method and the ``cheap translation'' baseline of \citep{D17-1268}. The size of the dictionaries used by both approaches are given in the right-most column. Using our model on their translated data from English outperforms the baseline scores produced by their models over all languages, a testament to the strength of our neural CRF baseline. The results produced by our model on their data indicate that our approach is effective, as we manage to outperform their approaches on all three languages using much smaller dictionaries and even without dictionaries. 
Also, we see that self-attention is effective when applied on their data, which also does not carry the correct word order.


\subsubsection{Why Does Translation Work Better?}
\begin{table*}[h!]
  \begin{center}
    \begin{tabular}{l|l|l|l}
      {Model} & Spanish & Dutch & German \\
      \hline
      Common space & $65.40 \pm 1.22$& $66.15 \pm 1.62$ & $43.73 \pm 0.94$\\
      Replace & $68.21 \pm 1.22$ & $69.37 \pm 1.33$ & $48.59 \pm 1.21$\\
      Translation & $\mathbf{69.21} \pm 0.95$ & $\mathbf{69.39} \pm 1.21$ & $\mathbf{53.94} \pm 0.66$\\
    \end{tabular}
    \caption{Comparison of different ways of using bilingual word embeddings, within our method (NER $F_1$).}
    \label{tab:table5}
  \end{center}
  \vspace{-1em}
\end{table*}

\begin{table*}[h!]
  \begin{center}\resizebox{\textwidth}{!}{
    \begin{tabular}{rl|l|l}
      \multicolumn{2}{l}{Model} & Uyghur Unsequestered Set& Extra Resources\\
      \hline
      $^{\ast\dagger}$&\citet{D17-1268} & $\mathbf{51.32}$ & Wikipedia, 100K dict.\\
      $^{\ast}$&\citet{D17-1268} (only Eng.~data)& $27.20$ & Wikipedia, 100K dict.\\
      \hline
      &BWET & $25.73 \pm 0.89$ &5K dict.\\
      & BWET + self-att. & $26.38 \pm 0.34$ &5K dict.\\
      $^{\ast}$&BWET on data from \citet{D17-1268}  & $30.20 \pm 0.98$ & Wikipedia, 100K dict.\\
      $^{\ast}$&BWET + self-att. on data from \citet{D17-1268}  & $30.68 \pm 0.45$ & Wikipedia, 100K dict.\\
      \hline
      $^{\ast}$&Combined  (see text) & $31.61 \pm 0.46$ & Wikipedia, 100K dict., 5K dict.\\
      $^{\ast}$&Combined + self-att. & $\mathbf{32.09} \pm 0.61$ & Wikipedia, 100K dict., 5K dict.\\
    \end{tabular}}
    \caption{NER $F_1$ scores on Uyghur.  $^\ast$Approaches using language-specific features and resources (``Wikipedia'' means Wikipedia is used not as a monolingual corpus, but to provide external knowledge). $^\dagger$Approaches that transfer from multiple languages and use language-specific techniques.}
    \label{tab:table6}
  \end{center}
  \vspace{-1em}
\end{table*}

In this section, we study the effects of different ways of using bilingual word embeddings and the resulting induced translations. 
As we pointed out previously, finding translations has two advantages: (1) the model can be trained on the exact points from the target embedding space, and (2) the model has access to the target language's original character sequences.
Here, we conduct ablation studies over these two variables. Specifically, we consider the following three variants.\footnote{In this study, we use GloVe for learning bilingual embeddings and word translations instead of fastText.}
\begin{itemize}[leftmargin=1.0em,topsep=0.5em,itemsep=0em]
\item \textbf{Common space}
This is the most common setting for using bilingual word embeddings, and has recently been applied in NER~\citep{DBLP:journals/corr/NiDF17}. 
In short, the source and target word embeddings are cast into a common space, namely $X'=XV$ and $Y'=YU$, and the model is trained with the source side embedding and the source character sequence, and directly applied on the target side.

\item \textbf{Replace}
In this setting, we replace each original word embedding $x_i$ with its nearest neighbor $y_i$ in the common space but do not perform translation. This way, the model will be trained with target word embeddings and source-side character sequences.
\item \textbf{Translation}
This is our proposed approach, where the model is trained on both  exact points in the target space and  target language character sequences.
\end{itemize}

The three variants are compared in Table~\ref{tab:table5}.
The ``common space'' variant performs the worst by a large margin, confirming our hypothesis that discrepancy between the two embedding spaces harms the model's  ability to generalize. 
From the comparison between the ``replace'' and ``translation,'' we observe that having access to the target language's character sequence helps  performance, especially for German, perhaps due in part to its capitalization patterns, which differ from English. In this case, we have to lower-case all the words for character inputs in order to prevent the model from overfitting the English capitalization pattern.

\subsection{Case Study: Uyghur}

In this section, we directly apply our approach to Uyghur, a truly low-resource language with very limited monolingual and parallel resources. We test our model on 199 annotated evaluation documents from the DARPA LORELEI program (the ``unsequestered set'') and compare with previously reported results in the cross-lingual setting by~\citet{D17-1268}. Similar to our previous experiments, we transfer from English, use fastText embeddings trained on Common Crawl and Wikipedia\footnote{\url{https://github.com/facebookresearch/fastText/blob/master/docs/crawl-vectors.md}} and a provided dictionary to perform translation, and use GloVe trained on a monolingual corpus that has 30 million tokens to perform NER. Results are presented in Table~\ref{tab:table6}.

Our method performs competitively, considering that we use a much smaller dictionary than~\citet{D17-1268} and no knowledge from Wikipedia  in Uyghur. Our best results come from a combined approach:  using word embeddings to translate words that are not covered by ~\citet{D17-1268}'s dictionary (last line of Table~\ref{tab:table6}). Note that for the CoNLL languages, ~\citet{D17-1268} used Wikipedia for the Wikifier features~\citep{TsaiMaRo16}, while for Uyghur they used it for translating named entities, which is crucial for low-resource languages when some named entities are not covered by the dictionary or the translation is not reliable. We suspect that the unreliable translation of named entities is the major reason why our method alone performs worse but performs better when combined with their data that has access to higher quality translations of named entities.

The table omits results using adversarial learning and identical character strings, as both failed ($F_1$ scores around 10).  We attribute these failures to the low quality of Uyghur word embeddings and the fact that the two languages are distant. Also, Uyghur is mainly written in Arabic script, making the identical character method inappropriate. Overall, this reveals a practical challenge for multilingual embedding methods, where the underlying distributions of the text in the two languages are divergent.

\section{Related Work}
\label{rw}




\textbf{Cross-Lingual Learning} Cross-lingual learning approaches can be loosely classified into two categories: annotation projection and language-independent transfer.

Annotation projection methods create training data by using parallel corpora to project annotations from the source to the target language. Such approaches have been applied to many tasks under the cross-lingual setting, such as POS tagging~\citep{H01-1035,P11-1061,DBLP:journals/tacl/TackstromDPMN13,DBLP:conf/conll/FangC16}, mention detection~\citep{Zitouni:2008:MDC:1613715.1613789} and parsing~\citep{hwa2005bootstrapping,McDonald:2011:MTD:2145432.2145440}.

Language independent transfer-based approaches build models using language independent and delexicalized features. For instance, ~\citet{P15-2064} transfers word cluster and gazetteer features through the use of comparable copora. ~\citet{TsaiMaRo16} links words to Wikipedia entries and uses the entry category as features to train language independent NER models.
Recently, \citet{DBLP:journals/corr/NiDF17} propose to project word embeddings into a common space as language independent features. These approaches utilize such features by training a model on the source language and directly applying it to the target language.

Another way of performing language independent transfer resorts to multi-task learning, where a model is trained jointly across different languages by sharing parameters to allow for knowledge transfer~\citep{ammar:2016:tacl, DBLP:journals/corr/YangSC17,I17-2016,lin2018multi}. However, such approaches usually require some amounts of training data in the target language for bootstrapping, which is different from our unsupervised approach that requires no labeled resources in the target language.

\paragraph{Bilingual Word Embeddings}
There have been two general paradigms in obtaining bilingual word vectors besides using dictionaries: through parallel corpora and through joint training. Approaches based on parallel corpora usually learn bilingual word embeddings that can produce similar representations for aligned sentences ~\citep{Hermann2014MultilingualMF,ap2014autoencoder}. Jointly-trained models combine the common monolingual training objective with a cross-lingual training objective that often comes from parallel corpus ~\citep{zou2013bilingual,gouws2015bilbowa}. Recently, unsupervised approaches also have been used to align two sets of word embeddings by learning a mapping through adversarial learning or self-learning~\citep{zhang2017adversarial,artetxe2017learning,lample2018word}.

\section{Conclusion}
\label{ccls}


In this paper, we propose two methods to tackle the cross-lingual NER problem under the unsupervised transfer setting. To address the challenge of lexical mapping, we  find translations of words in a shared embedding space built from a seed lexicon. To alleviate word order divergence across languages, we add a self-attention mechanism to our neural architecture. With these methods combined, we are able to achieve state-of-the-art or competitive results on commonly tested languages under a cross-lingual setting, with  lower resource requirements than past approaches. We also evaluate the challenges of applying these methods to an extremely low-resource language, Uyghur.
\section*{Acknowledgments}
\label{acknowledgements}
We thank Stephen Mayhew for sharing the data, and Zihang Dai for meaningful discussion.\\
This research was sponsored by Defense Advanced Research Projects Agency Information Innovation Office (I2O) under the Low Resource Languages for Emergent Incidents (LORELEI) program, issued by DARPA/I2O under Contract No. HR0011-15-C0114. The views and conclusions contained in this document are those of the authors and should not be interpreted as representing the official policies, either expressed or implied, of the U.S.~government. The U.S.~government is authorized to reproduce and distribute reprints for government purposes notwithstanding any copyright notation here on.

\bibliography{emnlp2018}
\bibliographystyle{acl_natbib_nourl}

\end{document}